# Comparative Analysis of Machine Learning Approaches for Bone Age Assessment: A Comprehensive Study on Three Distinct Models

Nandavardhan Radhakrishnan
Student
Networking and Communications
*SRM Institute of Science & Technology*
Chennai, India
nr3648@srmist.edu.in

Somanathan Raghuraman
Student
Networking and Communications
*SRM Institute of Science & Technology*
Chennai, India
sr0451@srmist.edu.in

Vikram Suresh
Student
Networking and Communications
*SRM Institute of Science & Technology*
Chennai, India
vs7552@srmist.edu.in

Dr.P.Savaridassan
Assistant Professor
Networking and Communications
*SRM Institute of Science & Technology*
Chennai, India
savaridp@srmist.edu.in

*Abstract*—Background: Radiologists and doctors make use of X-ray images of the non-dominant hands of children and infants to assess the possibility of genetic conditions and growth abnormalities. This is done by assessing the difference between the actual extent of growth found using the X-rays and the chronological age of the subject. The assessment was done conventionally using The Greulich–Pyle (GP) or Tanner–Whitehouse (TW) approach. These approaches require a high level of expertise and may often lead to observer bias. Hence, to automate the process of assessing the X-rays, and to increase its accuracy and efficiency, several machine learning models have been developed. These machine-learning models have several differences in their accuracy and efficiencies, leading to an unclear choice for the suitable model depending on their needs and available resources.

Methods: In this study, we have analyzed the 3 most widely used models for the automation of bone age prediction, which are the Xception model, VGG model and CNN model. These models were trained on the preprocessed dataset and the accuracy was measured using the MAE in terms of months for each model. Using this, the comparison between the models was done.

Results: The 3 models, Xception, VGG, and CNN models have been tested for accuracy and other relevant factors. The Xception model is the most accurate among the three models evaluated with an MAE (months) of 12.6. It contains 20 million parameters. It is the largest model amongst the three models. The VGG model produced an MAE (months) of 34.9. The model has 17 million parameters. The VGG model is relatively the least accurate but is smaller in size when compared with the Xception model. The CNN model contains the least number of parameters with 3 million, making it the smallest in size, and has an MAE (months) of 22.6 achieving an accuracy higher than the VGG model but lower than that of the Xception model.

Conclusions: Amongst the three models tested, The Xception model has been found to produce results with the highest level of accuracy. However, owing to its large size and complexity, it may not be as easy to implement as the other models and is more resource intensive. The CNN model is the most lightweight among the three and hence can be used on a wider range of devices although accuracy is compromised when compared to the Xception model. The VGG model is the easiest to implement, however lacks accuracy.

## I. BACKGROUND

X-ray images of the non-dominant hands of children and infants are examined by doctors and radiologists to assess the possibility of genetic conditions and growth abnormalities. This analysis is done by assessing the discrepancy between the actual extent of bone growth found using the X-rays and the chronological age of the subject [1].

The Greulich–Pyle (GP) approach uses the skeletal structure of the wrist to evaluate the bone age of children. This approach is predominantly used in children of European descent. This approach is preferred among radiologists owing to its ease of implementation although it is not preferred when analyzing the bones of children with certain medical conditions [2].

The Tanner–Whitehouse (TW) methodology is a medical methodology that is used to determine bone age particularly amongst individuals during the adolescent growth period by evaluating skeletal maturity. It is a more quantitative and detailed methodology to determine skeletal maturity [3].

Both conventional approaches are subject to inter-observer variability and potentially lead to misdiagnosis. Moreover, the application and implementation of this model may require a higher level of expertise.

Hence, to reduce the effect of such human factors and biases in the assessment of the bone age, we utilize several machine learning models [4]. These machine learning models increase accuracy and automate the whole process, thereby increasing efficiency.

Several machine learning models have been developed to tackle this specific problem. These models vary greatly in their approach and their accuracy and efficiency.
Prior research on this topic focuses on the performance of singular models and their results. However, a direct comparison between separate models for this specific problem statement has not been explored in detail yet.

Our research highlights the performance of various models and their relative indicators.

This paper aims to give medical institutions and radiologists a well-informed understanding as to which automated bone age assessment model would suit their available infrastructure and requirements.

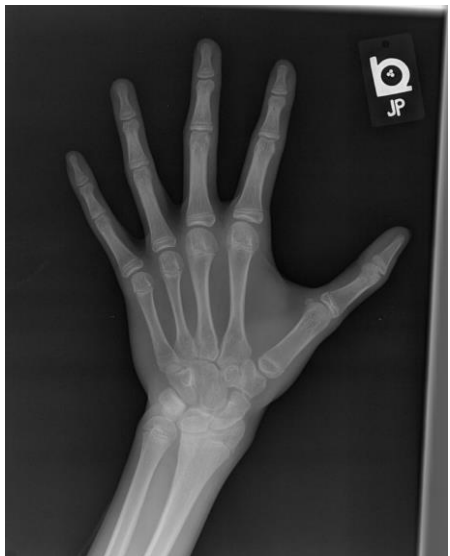

**Figure 1.** Sample X-Ray image in the dataset

## II. METHODS

The machine learning models have been implemented using Keras, an open-source neural networks application programming interface (API) built for developing and training deep learning models.

The dataset obtained from Kaggle, 'RSNA Bone Age+Anatomical ROIS' contains 12611 images of training data, with their corresponding bone age, and similarly contains 1425 validation images and their respective bone age.

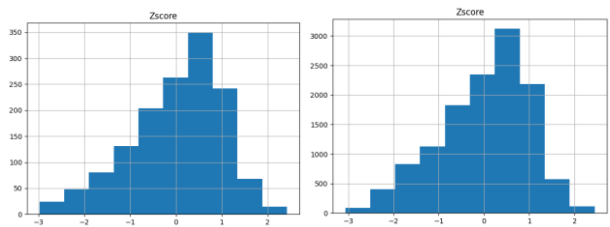

**Figure 3.** Zscore of validation and training images respectively

The preprocessing of the dataset for both the VGG and CNN model was done by converting the images to greyscale and resizing the images to a resolution of 256x256 with the help of image data generators in Keras.

The preprocessing for the Xception model differs such that the resolution has been further scaled down to 128x128 in order to reduce the memory requirements for training. This was followed by using the inbuilt Keras function to normalize the images.

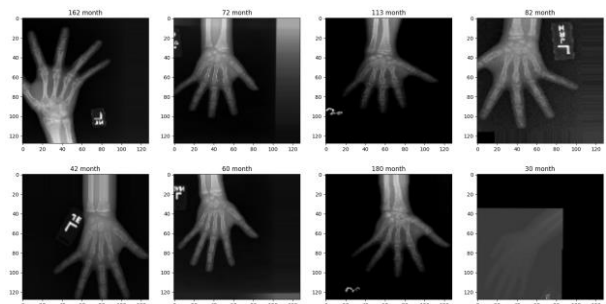

**Figure 2.** Training images after preprocessing

CNN Model

Convolutional Neural Networks (CNNs) are a collection of deep neural networks that are used for image recognition and computer vision [5-7]. The CNN model has been developed specifically for accommodating images of the resolution 256x256. The model consists of alternate layers of Conv2D layers, BatchNormalization and MaxPooling2D layers along with a dropout layer to reduce overfitting. This model has 4 Conv2D layers and the output is a single dense node which provides the predicted bone age in terms of months.

| Layer | Output Shape | Parameters |
|---|---|---|
| InputLayer | (None, 256, 256, 1) | 0 |
| Conv2D | (None, 256, 256, 32) | 320 |
| BatchNormalization | (None, 256, 256, 32) | 128 |
| MaxPooling2D | (None, 128, 128, 32) | 0 |
| Conv2D | (None, 128, 128, 64) | 18496 |
| BatchNormalization | (None, 128, 128, 64) | 256 |
| MaxPooling2D | (None, 64, 64, 64) | 0 |
| Dropout | (None, 64, 64, 64) | 0 |
| Conv2D | (None, 64, 64, 128) | 73856 |
| BatchNormalization | (None, 64, 64, 128) | 512 |
| MaxPooling2D | (None, 32, 32, 128) | 0 |
| Dropout | (None, 32, 32, 128) | 0 |
| Conv2D | (None, 32, 32, 256) | 295168 |
| BatchNormalization | (None, 32, 32, 256) | 1024 |
| MaxPooling2D | (None, 16, 16, 256) | 0 |
| Dropout | (None, 16, 16, 256) | 0 |
| Flatten | (None, 65536) | 0 |
| Dense | (None, 50) | 3276850 |
| Dense | (None, 10) | 510 |
| Dense | (None, 1) | 11 |

## VGG Model

The Visual Geometry Group (VGG) model is a family of deep convolutional neural networks (CNN) used in image classification. It is a general-purpose model and has been modified to suit our requirements. This allows it to be seamlessly integrated into our project [8,9].

| Layer | Output Shape | Parameters |
|---|---|---|
| InputLayer | (None, 256, 256,1) | 0 |
| Conv2D | (None, 256, 256, 64) | 640 |
| Conv2D | (None, 256, 256, 64) | 36928 |
| MaxPooIing2D | (None, 128, 128,64 | 0 |
| Conv2D | (None, 128, 128, 128) | 73856 |
| Conv2D | (None, 128, 128, 128) | 147584 |
| MaxPooIing2D | (None, 64, 64, 128) | 0 |
| Conv2D | (None, 64, 64, 256) | 295168 |
| Conv2D | (None, 64, 64, 256) | 590080 |
| Conv2D | (None, 64, 64, 256) | 590080 |
| MaxPooIing2D | (None, 32, 32, 256) | 0 |
| Conv2D | (None, 32, 32, 512) | 1180160 |
| Conv2D | (None, 32, 32, 512) | 2359808 |
| Conv2D | (None, 32, 32, 512) | 2359808 |
| MaxPooIing2D | (None, 16, 16, 512) | 0 |
| Conv2D | (None, 16, 16, 512) | 2359808 |
| Conv2D | (None, 16, 16, 512) | 2359808 |
| Conv2D | (None, 16, 16, 512) | 2359808 |
| MaxPooIing2D | (None, 8, 8, 512) | 0 |
| Flatten | (None, 32768) | 0 |
| Dense | (None, 100) | 3276900 |
| Dense | (None, 10) | 1010 |
| Dense | (None, 1) | 11 |

## Xception Model

The Xception model is a deep convolutional neural network (CNN) architecture. It makes use of a novel approach called the depthwise separable convolutions. The purpose of these convolutions is to capture the spatial and channel-wise dependencies in the data while reducing the number of parameters compared to traditional convolutions. Depthwise Separable Convolutions perform standard convolution in two steps: Depthwise convolution which applies a single filter to each channel and Pointwise convolution which are 1x1 convolution to combine the output of multiple channels [10-12].

The Xception model consists of 71 layers which leads to better learning of complex features present in images. The Xception model features residual connections which work similarly to shortcuts that work by bypassing certain parts of the model, this prevents diminishing gradients. Furthermore, the architecture of the Xception model consists of 3 parts. The entry flow is responsible for extracting high-level features. The middle flow, comprising multiple blocks of depthwise separable convolutions arranged linearly. Finally, the exit flow, characterized by pooling and dense layers, is designed to prepare the extracted features for regression.

## III. RESULTS

The 3 machine learning models, Xception, VGG, and CNN models have been tested for accuracy and other relevant factors. The accuracy of the models has been uniformly measured using the Mean Absolute Error in terms of months.

The Xception model demonstrated higher accuracy when compared to the VGG and CNN models.
The Xception model achieved an MAE of 12.6 months. It utilizes [20,881,981] parameters and consists of 71 layers, thereby making it difficult to implement.

Upon examining the CNN model, we obtained an MAE of 22.6 months. This implies that the CNN model is comparatively more accurate than the VGG model but less accurate than the Xception model. The CNN model uses the least number of parameters [3,667,131] which makes it the smallest in size. The CNN model is easier to implement than the Xception model due to its simple architecture but lacking in comparison to the VGG model in terms of implementation. Hence it is preferred in situations where there is a hardware limitation.

On the other hand, the VGG model produced an MAE of 34.9 months. This suggests that the accuracy obtained from using the VGG model is less than the accuracy produced by the Xception model. The VGG model contains [17,991,457] parameters making it smaller in size, and it being a standardized architecture, makes it easier to implement relative to the other two models.

## IV. CONCLUSIONS

The comparison of the three machine learning models has been done based on the selected three parameters, including the accuracy measured using MAE in terms of months, size measured using the number of parameters and the ease of implementation of the machine learning model.

The comparison and analysis of the accuracy and usability of the three bone age assessment models have been performed in this paper. These models had been developed as an alternative to the traditional methods of bone age assessment for growth anomaly in infants and children. The models that have been studied in this paper are the Xception model, the VGG model, and the CNN model.

This paper proposes the differences in these models by means of accuracy using the Mean Absolute Error (MAE) in terms of months as the metric. The size and implementation aspects of these three models have also been considered to provide us with a better understanding of the proper requirements needed to train and deploy each model. The most accurate model among the three, the Xception model, has been tested to have an MAE of 12.6 months. This places it as the model with the highest accuracy among the three. Although it has the highest accuracy, it also comes with some limitations.

The Xception model uses [20,881,981] parameters making it the most complicated among the three. This may be harder to run on devices with lower hardware specifications and is not as efficient and lightweight as the VGG and CNN models.

The advantage of the VGG model being a general-purpose model means it is easier to implement. This makes it preferable for those who don't have much expertise in the subject of image classification. The number of parameters in the VGG model is [17,991,457]. This makes it more lightweight than the Xception model. Where the VGG model lacks is in terms of accuracy, scoring an MAE of 34.9488 which means it has the highest mean absolute error compared to the other two, making it the least in terms of accuracy. Although this model is well-suited to those who don't have much expertise on the subject and want something quick and convenient to implement, the other two models promise a higher level of accuracy which would be deemed necessary for the accurate prediction of growth anomalies in individuals.

The CNN model was measured to have an MAE of 22.6, which is more accurate when compared to the VGG model but not as accurate as Xception model. The CNN model has [3,667,131] parameters making it the smallest among the three models. While not as easy to implement as the VGG model, would be viewed as an alternative to those who can't run the Xception model with their limited infrastructure.

This paper concludes that the Xception model offers the highest accuracy while being the most resource intensive among the three. The CNN model offers a lower accuracy when compared to the Xception model, but tradeoffs in accuracy are offset in terms of resource efficiency, as it has the least number of parameters. Thus, making it usable for a wider range of devices. The VGG model, although the easiest to implement, lacks in comparison to the CNN model in both accuracy and size.

## V. CITATIONS